\documentclass{article}

\usepackage[preprint]{neurips_2025}

\usepackage[utf8]{inputenc} 
\usepackage[T1]{fontenc}    
\usepackage{hyperref}       
\usepackage{url}            
\usepackage{booktabs}       
\usepackage{amsfonts}       
\usepackage{nicefrac}       
\usepackage{microtype} 
\usepackage{tikz}
\usepackage{tabularx}
\usepackage{xcolor}         
\usepackage{lipsum}
\usepackage{graphicx}
\usepackage{amssymb}  
\usepackage{pifont}   
\usepackage{algorithm}
\usepackage{algpseudocode}
\usepackage{amsmath}
\usepackage{multirow, array}
\usepackage{tabularx}
\usepackage{wrapfig}
\usepackage{enumitem}
\usepackage{siunitx}
\usepackage{graphicx}  
\usepackage{pgfplots} 
\pgfplotsset{compat=1.17} 
\usepackage{newfloat}
\usepackage{listings}
\usepackage{threeparttable}
\usepackage{subfigure}

\usepackage{varwidth}

\DeclareMathOperator*{\argmax}{arg\,max}

\title{Diffusion with a Linguistic Compass: Steering the Generation of Clinically Plausible Future sMRI Representations for Early MCI Conversion Prediction}

\author{%
  Zhihao Tang$^1$, \enspace Chaozhuo Li$^1$, \enspace Litian Zhang$^1$, \enspace Xi Zhang$^1$\thanks{Corresponding author.}\\ 
  $^1$Key Laboratory of Trustworthy Distributed Computing and Service (MoE), \\
  Beijing University of Posts and Telecommunications, \\
  Beijing 100876, China \\
  \texttt{\{lichaozhuo, innerone, zhangx\}@bupt.edu.cn, litianzhang@buaa.edu.cn}
}

\begin{document}

\maketitle

\begin{abstract}
  Early prediction of Mild Cognitive Impairment (MCI) conversion is hampered by a trade-off between immediacy—making fast predictions from a single baseline sMRI—and accuracy—leveraging longitudinal scans to capture disease progression. We propose MCI-Diff, a diffusion-based framework that synthesizes clinically plausible future sMRI representations directly from baseline data, achieving both real-time risk assessment and high predictive performance. First, a multi-task sequence reconstruction strategy trains a shared denoising network on interpolation and extrapolation tasks to handle irregular follow-up sampling and learn robust latent trajectories. Second, an LLM-driven “linguistic compass” is introduced for clinical plausibility sampling: generated feature candidates are quantized, tokenized, and scored by a fine-tuned language model conditioned on expected structural biomarkers, guiding autoregressive generation toward realistic disease patterns. Experiments on ADNI and AIBL cohorts show that MCI-Diff outperforms state-of-the-art baselines, improving early conversion accuracy by 5–12\%.
\end{abstract}

\section{Introduction}

The prediction of MCI (Mild Cognitive Impairment) conversion aims to forecast the clinical trajectory of individuals diagnosed with MCI~\cite{2006mci, ADsurvey, Fan2022CancerSP, so5, so6}. MCI patients typically exhibit two progression patterns: a subset may deteriorate into progressive MCI (pMCI), while others remain clinically stable (sMCI)~\cite{mci2}. Accurate classification of patients into pMCI or sMCI subgroups has significant implications for personalized treatment planning and improved clinical trial stratification~\cite{mci, augdiff, deepmcisurvey}.

Conventional models for predicting MCI conversion primarily rely on Structural Magnetic Resonance Imaging (sMRI) data, as sMRI captures nuanced patterns of neural alterations associated with brain structural changes. 
Existing sMRI-based models can be broadly categorized into two types: cross-sectional and longitudinal methods~\cite{deepmcisurvey, so1, so2, so3, so4}. 
Cross-sectional models generate predictions from a single baseline sMRI scan obtained at the initial timepoint~\cite{MPS-FFA, HFCN, DA-MIL, guan2021domain, so7}, whereas longitudinal approaches utilize serial sMRI scans acquired at multiple follow-up intervals to model temporal dynamics in brain morphology \cite{2023vggts, wang2018temporal, RNNMCI, mri3, mri4}.  

\begin{figure}[t]
  \centering
\includegraphics[width=\linewidth]{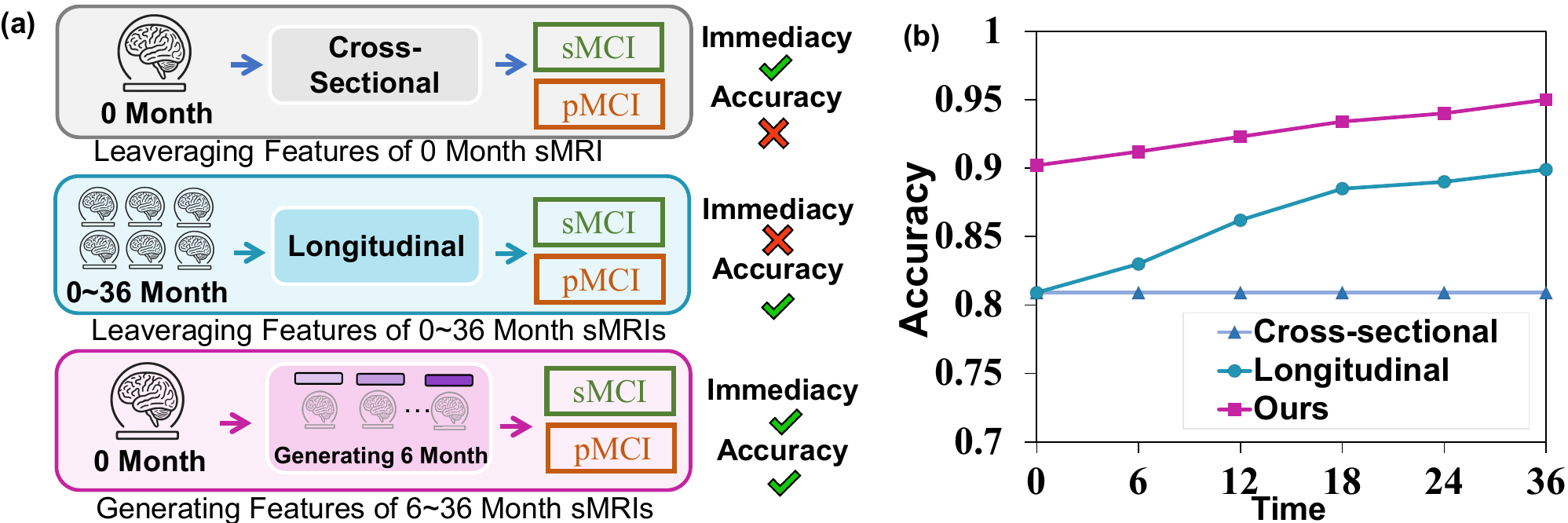}
\vspace{-5mm}
    \caption{ 
    Comparison of MCI Conversion Prediction Models and Performance. \textbf{(a)} Schematics illustrate the input sMRI data for cross-sectional (baseline), longitudinal (0 to 36 months), and our generated longitudinal-like feature approaches, highlighting their implications for prediction immediacy and accuracy. \textbf{(b)} Accuracy of these three approaches over a 36-month prediction window.
    }
   \label{fig:intro}
   \vspace{-5mm}
\end{figure}

Despite advancements in existing models, a fundamental trade-off persists: the inherent trade-off between prediction immediacy and accuracy.
Immediacy quantifies the latency between data acquisition and actionable predictions, while accuracy reflects the effectiveness in capturing ground-truth outcomes. 
As illustrated in Fig.~\ref{fig:intro}(a), cross-sectional methods leverage only a baseline sMRI scan (e.g., at 0 months), enabling rapid predictions (high immediacy).
However, their reliance on a single timepoint restricts access to temporal dynamics, resulting in suboptimal accuracy due to limited anatomical context and absence of disease progression signals. 
For instance, Fig.~\ref{fig:intro}(b) demonstrates that HFCN~\cite{HFCN} trained solely on 0-month sMRIs underperforms its longitudinal counterpart (integrating sMRIs from 0–36 months) on the ADNI dataset~\cite{Vaswani2017AttentionIA, so8}. 
Conversely, longitudinal methods exploit serial imaging data to capture temporal trajectories, thereby enhancing accuracy. Yet, their dependency on historical scans (e.g., 36-month follow-ups) inherently compromises immediacy, delaying predictions until sufficient longitudinal data is acquired.

Here comes a novel question:  \textit{Can we leverage the merits of both approaches to achieve a win-win balance between immediacy and accuracy?} 
In contrast to cross-sectional models (which lack temporal dynamics) and longitudinal models (which suffer from delayed predictions), our goal is to learn informative longitudinal features derived from baseline sMRI scans as complementary data.
As illustrated in Fig.~\ref{fig:intro}(a), by generating latent longitudinal trajectories from a single baseline scan, our proposal enables accurate predictions at the earliest time point while eliminating the latency inherent in longitudinal data acquisition.  
This generation-based paradigm, relying exclusively on baseline sMRI, is expected to deliver predictive performance comparable to longitudinal methods while retaining the critical advantage of real-time risk assessment. 

However, effectively generating realistic longitudinal sMRI sequences that accurately reflect the subtle and heterogeneous progression in MCI patients remains a significant challenge.
First, accurately modeling the complex distribution of longitudinal sMRI sequences is difficult for existing generative models. For example, GAN-based models~\cite{goodfellow2020generative} often face instability issues, while VAE-based models may suffer from posterior collapse~\cite{VAE}.
Second, the high dimensionality and intricate spatial correlations in sMRI data challenge traditional generative models due to their limited capacity for high-dimensional representation~\cite{frisoni2010clinical}.
Such limitations restrict the ability of conventional generative models to capture nuanced MCI progression patterns.

To address these challenges, we propose a diffusion-based framework for generating realistic longitudinal sMRI sequences. 
Leveraging its stable denoising process to learn complex data distributions~\cite{croitoru2023diffusion}, the diffusion model could effectively capture the subtle and heterogeneous progression within sequences.
To mitigate computational complexity, we focus on synthesizing low-dimensional feature representations of future scans. 
However, directly applying vanilla denoising diffusion models to our task is undesirable.  
First, the irregular temporal sampling inherent in longitudinal MCI studies conflicts with the uniform time-step assumption of standard diffusion architectures. This mismatch impairs modeling of complex progression trajectories, particularly when handling missing intermediate or terminal time points.  
Second, autoregressive generation risks compounding errors over time. Even minor deviations from actual neurodegeneration patterns which exacerbated by uneven time intervals may snowball across predictions, reducing conversion accuracy. 

In this paper, we propose MCI-Diffusion (\textbf{MCI-Diff}), a novel framework for generating clinically plausible future sMRI representations to predict early MCI conversion. 
MCI-Diff employs two key strategies: \textbf{Multi-task Sequence Reconstruction Training} and \textbf{LLM-driven Clinical Plausibility Sampling}. 
The former tackles the irregular temporal sampling by training a shared denoising network conditioned on task-specific information to perform both interpolation and extrapolation, effectively learning to handle non-uniform time steps.
The latter mitigates error compounding in autoregressive future sMRI generation by using an LLM to iteratively score and select plausible outputs based on expected progression, thereby guiding the generation towards clinically coherent trajectories.
By leveraging baseline sMRIs to predict future representations, MCI-Diff enables early and accurate MCI conversion prediction. 
We evaluate MCI-Diff on two real-world datasets, demonstrating its effectiveness in generating longitudinal MRI sequences and improving early prediction accuracy. 

In summary, our contributions are as follows: First, to our knowledge, this work is the first to leverage diffusion models for synthesizing future MRI representations specifically aimed at enhancing early MCI conversion prediction. Second, we introduce MCI-Diff, a novel framework employing multi-task sequence reconstruction training and LLM-driven clinical plausibility sampling to address the challenges of irregular temporal sampling and error compounding in generating future sMRI representations. Finally, extensive experiments on two widely adopted real-world datasets demonstrate that MCI-Diff generates clinically plausible longitudinal MRI representations and achieves improved accuracy in early MCI conversion prediction compared to existing methods.

\section{Problem Setup}

\subsection{Definition of Longitudinal sMRI Data and MCI Conversion Outcome}
Consider a cohort of MCI individuals $\mathcal{P} = \{p_1, \ldots, p_{|\mathcal{P}|}\}$, where each individual $p$ has a baseline sMRI $X_0^{(p)} \in \mathcal{X}$ and a conversion label $Y^{(p)} \in \{\mathrm{pMCI, sMCI}\}$.
In the context of longitudinal analysis (even though our goal is to avoid relying on it directly for prediction), each MCI patient $p$ is scheduled for sMRI scans at fixed time points $\mathcal{T} = \{0,6,12,18,24,36\}\,\text{months}$ (with 0 denoting baseline), but some patients may miss one or more visits.
We denote the actually observed scans for patient $p$ as $\{\,X_{\tau}^{(p)} : \tau \in \mathcal{T}_p\}$, where $X_{\tau}^{(p)}\in\mathcal{X}$ is the sMRI at time $\tau$, and $\mathcal{T}_p \subseteq \mathcal{T}$ is the subset of time points at which patient $p$ has available data. For notational convenience, $\tau$ here represents the position of the scan within the complete sequence $\mathcal{T}$, rather than the specific time in months.

\subsection{Definition of Early MCI Conversion Prediction via Future sMRI Generation}
The standard cross-sectional MCI conversion prediction task aims to predict the outcome $Y^{(p)}$ based solely on the baseline sMRI $X_0^{(p)}$. Formally:
\begin{equation}
\hat{Y}^{(p)} = \argmax_{y \in \{\mathrm{pMCI, sMCI}\}} Pr(Y^{(p)} = y \mid X_0^{(p)}).
\label{equ_baseline_prediction}
\end{equation}

Our motivation lies in generating future sMRI representations in an autoregressive manner from the baseline scan to enhance prediction accuracy while maintaining timeliness. 
Let $f_g( \cdot \mid \theta)$ be our generative model (MCI-Diffusion) with parameters $\theta$. 
Given a baseline sMRI $X_0^{(p)}$ and a latent noise vector $z_0$ sampled from a distribution $p(z)$ (e.g., a normal distribution), $f_g( \cdot \mid \theta)$ generates a sequence of future sMRI representations in a step-wise autoregressive fashion in a latent feature space:
\begin{align}
\hat Z_\tau^{(p)} =
\begin{cases}
f_{g}\bigl(z_{0},\,\phi(X_{0}^{(p)}) \mid \theta\bigr), 
& \tau = 1, \\[6pt]
f_{g}\bigl(z_{\tau},\,\hat Z_{1:\tau-1}^{(p)},\,\phi(X_{0}^{(p)}) \mid \theta\bigr), 
& \tau = 2,\ldots,|\mathcal{T}|-1.
\end{cases}
\label{equ_autoregressive_generation}
\end{align}
where $\phi: \mathcal{X} \rightarrow \mathcal{Z}$ is a pretrained feature extractor that maps an sMRI scan to a feature vector in a latent space $\mathcal{Z}$, and $\hat{Z}_\tau^{(p)}$ represents the generated feature representation in position $\tau$ of the sMRI sequence, whose length is $|\mathcal{T}|-1$ because we aim to predict future MRI representations at the follow-up time points after the baseline scan (time 0), and $|\mathcal{T}|$ includes the baseline time point.

The early MCI conversion prediction using these generated future representations is formulated as:
\begin{equation}
\hat{Y}^{(p)} = \argmax_{y \in \{\text{pMCI, sMCI}\}} Pr(Y^{(p)} = y \mid \phi(X_0^{(p)}), \{\hat{Z}_1^{(p)}, \ldots, \hat{Z}_{|\mathcal{T}|-1}^{(p)}\}).
\label{equ_autoregressive_prediction_short}
\end{equation}
This prediction leverages both baseline sMRI representation and the autoregressively generated future representations from our model $f_g$ (MCI-Diff).

\section{Methodology}

\begin{figure}[ht]
  \centering
\includegraphics[width=\linewidth]{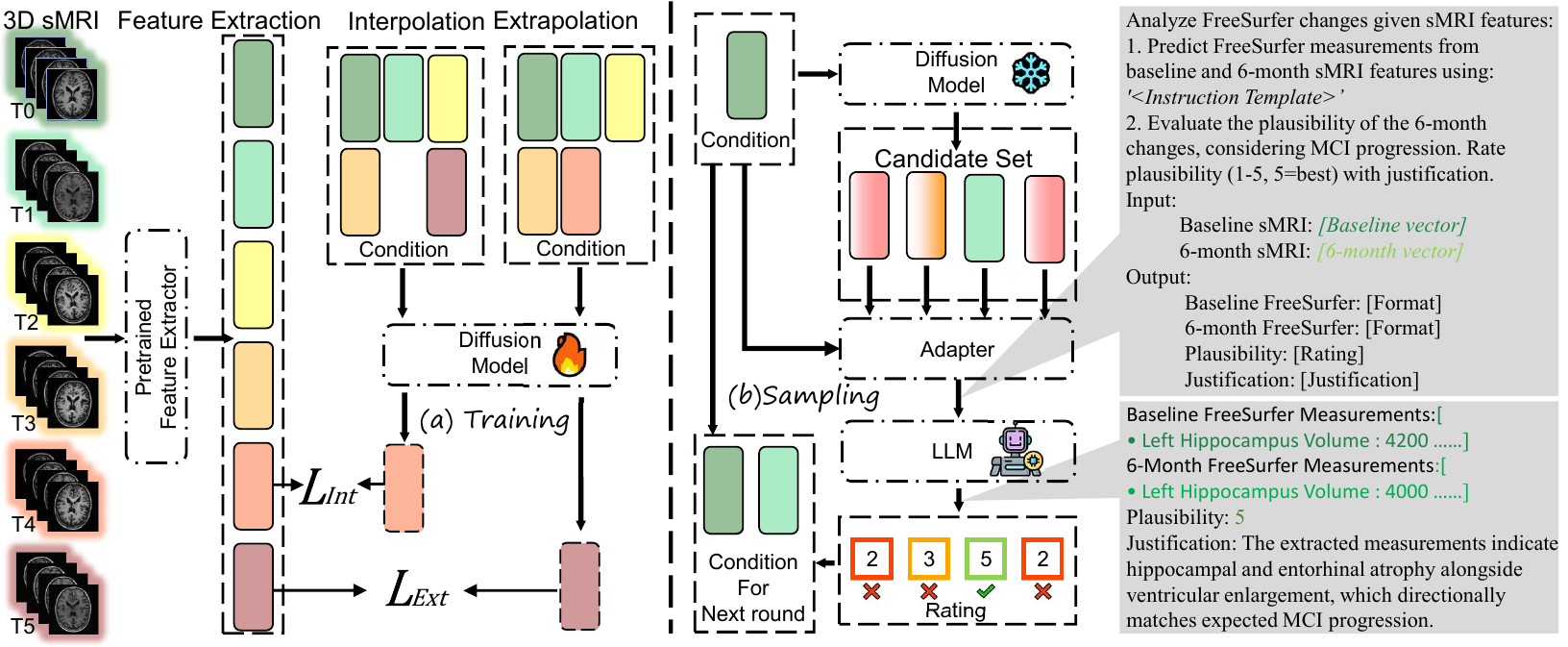}
\vspace{-5mm}
\caption{The overview of our MCI-Diff Framework.
  (a) \textbf{Multi-task Sequence Reconstruction Training:} A shared Diffusion Model learns to reconstruct sMRI representations at a specific time conditioned on representations at other times via interpolation ($\mathcal{L}_{\text{Int}}$) and extrapolation ($\mathcal{L}_{\text{Ext}}$), addressing irregular temporal sampling by imputing missing sMRI time points.
  (b) \textbf{LLM-driven Clinical Plausibility Sampling:} The Diffusion Model generates future sMRI feature candidates, which an Adapter prepares for LLM scoring based on MCI progression consistency. The LLM selects the most plausible candidate at each step to guide subsequent generation, mitigating error compounding.}
   \label{fig:method}
\end{figure}
\vspace{-3mm}
MCI-Diffusion (Fig.~\ref{fig:method}) comprises two stages: \textbf{Multi-task Sequence Reconstruction Training} for handling irregular temporal sampling, and \textbf{LLM-driven Clinical Plausibility Sampling} for generating clinically plausible future sMRI representations.

\subsection{Multi-task Sequence Reconstruction Training}

This section first introduces the architecture and inputs of the denoiser, the core component of our diffusion model. Then, it elaborates on the Interpolation task (leveraging existing data) and the Extrapolation task (generating future data to address data sparsity). Finally, it presents a progressive training schedule that alternates between these two tasks to enhance model robustness and generalization, effectively handling the irregular sampling issue.

\noindent \textbf{3.1.1 Denoiser Architecture.} Our denoiser models the temporal evolution of sMRI features for MCI conversion prediction, explicitly handling irregular sampling. Specifically, for a given patient $p$, we consider a fixed-length sequence of $|\mathcal{T}|$ time points. For each temporal index $\tau \in \{0, \ldots, |\mathcal{T}|-1\}$, we have an sMRI feature representation $\mathbf{Z}_\tau^{(p)} \in \mathcal{Z}$ ( a placeholder if missing).

The denoiser receives input at each reverse diffusion step $\tau$ as a combination of: the sMRI representation $\mathbf{Z}_\tau^{(p)}$ (padded if masked), the positional embedding $\mathbf{P}_\tau \in \mathbb{R}^{\dim(\mathcal{Z})}$ encoding time $\tau$, and the mask embedding $\mathbf{M}_\tau \in \mathbb{R}^{\dim(\mathcal{Z})}$ indicating presence (0) or absence (1) of sMRI data.

These components are combined element-wise to form the input sequence of condition. Additionally, a target positional embedding $\mathbf{T}_i \in \mathbb{R}^{dim(\mathcal{Z})}$ is used to indicate the specific temporal index $i \in \{0, \ldots, |\mathcal{T}|-1\}$ for which the denoiser is predicting the feature. The conditional denoising diffusion probabilistic model~\cite{DDPM} operates in forward and reverse processes.

\textit{Forward Process:} Noise is gradually added to the target representation $\mathbf{Z}_i^{(p)}$ over $T$ steps:
\begin{equation}
q(\mathbf{Z}_{i, 1:T}^{(p)} | \mathbf{Z}_{i, 0}^{(p)}) = \prod_{t=1}^T q(\mathbf{Z}_{i, t}^{(p)} | \mathbf{Z}_{i, t-1}^{(p)}), \quad
q(\mathbf{Z}_{i, t}^{(p)} | \mathbf{Z}_{i, t-1}^{(p)}) = \mathcal{N}(\mathbf{Z}_{i, t}^{(p)}; \sqrt{1 - \beta_t} \mathbf{Z}_{i, t-1}^{(p)}, \beta_t \mathbf{I}),
\end{equation}
where $\mathbf{Z}_{i, t}^{(p)}$ is the noisy target representation at step $t$, and $\beta_{1:T}$ is the noise schedule.

\textit{Reverse Process:} Starting from a noisy target representation $\mathbf{Z}_{i, T}^{(p)}$, the denoiser $\epsilon_\theta$ aims to predict the noise at each step $t$, conditioned on $\mathbf{C}$ and the target positional embedding $\mathbf{T}_i$. Here, $\mathbf{C} = [\mathbf{c}_0, \mathbf{c}_1, ..., \mathbf{c}_{|\mathcal{T}|-1}]$ is a sequence where $\mathbf{c}_\tau = \mathbf{Z}_\tau^{(p)} + \mathbf{P}_\tau + \mathbf{M}_\tau$ if $\tau \neq i$, and $\mathbf{c}_i = \mathbf{Z}_{\text{masked}} + \mathbf{P}_i + \mathbf{M}_i$. $\mathbf{Z}_{\text{masked}}$ represents the masked sMRI representation at position $i$. The reverse process is defined as:
\begin{equation}
p_\theta(\mathbf{Z}_{i, t-1}^{(p)} | \mathbf{Z}_{i, t}^{(p)}, \mathbf{C}, \mathbf{T}_i, t) = \mathcal{N}(\mathbf{Z}_{i, t-1}^{(p)}; \mathbf{\mu}_\theta(\mathbf{Z}_{i, t}^{(p)}, \mathbf{C}, \mathbf{T}_i, t)),
\end{equation}
where $\mathbf{\mu}_\theta$ is the predicted mean. The training objective is to minimize the L2 loss between the predicted noise and the actual noise added during the forward process:
\begin{equation}
\mathcal{L} = \mathbb{E}_{\mathbf{\epsilon} \sim \mathcal{N}(0, 1), t} [||\mathbf{\epsilon} - \epsilon_\theta(\mathbf{x}_t, \mathbf{T}_i, t)||_2^2],
\end{equation}
where $\mathbf{\epsilon}$ is the noise and $\epsilon_\theta$ is the predicted noise.

\begin{algorithm}[t]
\caption{Multi-task Sequence Reconstruction Training}
\label{alg:progressive_training}
\textbf{Input:} Original training dataset $\mathcal{D}_{\text{orig}}$ of sMRI sequences, Denoiser model $\epsilon_\theta$, Max difficulty $D_{\text{max}}$
\begin{algorithmic}[1]
\State $\mathcal{D} \leftarrow$ Subset of $\mathcal{D}_{\text{orig}}$ containing only complete sequences
\State Initialize difficulty level $d = 1$
\While{$d \leq D_{\text{max}}$}
    \Comment{Interpolation Phase}
    \State Train $\epsilon_\theta$ on $\mathcal{D}$ with Interpolation (masking $d$ intermediate points)
    \State $\mathcal{D} \leftarrow \mathcal{D} \cup$ (Imputed sequences from $\mathcal{D}_{\text{orig}}$ with $d$ missing intermediate points)
    
    \Comment{Extrapolation Phase}
    \State Train $\epsilon_\theta$ on $\mathcal{D}$ with Extrapolation (masking $d$ final points)
    \State $\mathcal{D} \leftarrow \mathcal{D} \cup$ (Imputed sequences from $\mathcal{D}_{\text{orig}}$ with $d$ missing final points)
    
    \State $d = d + 1$
\EndWhile
\end{algorithmic}
\end{algorithm}

\noindent \textbf{3.1.2 Interpolation Task.} The Interpolation task serves two key purposes: initial denoiser training for intermediate feature reconstruction and subsequent data augmentation by imputing missing intermediate time points. MCI-Diffusion is trained to predict masked sMRI features at intermediate time points. For each complete sequence, we randomly mask an intermediate temporal index $i \in \{1, \ldots, |\mathcal{T}|-2\}$ and condition the denoiser $\epsilon_\theta$ on the remaining sequence $\mathbf{C}$ (with the $i$-th position masked) and a target positional embedding $\mathbf{T}_i$. The training objective is defined as:
\begin{equation}
L_{\text{Int}} = \mathbb{E}_{\mathbf{\epsilon} \sim \mathcal{N}(0, 1), t} \left[ ||\mathbf{\epsilon} - \epsilon_\theta(\mathbf{x}_t, \mathbf{C}, \mathbf{T}_i, t)||_2^2 \right], \quad i \in \{1, ..., |\mathcal{T}| - 2\}.
\end{equation}
where $\mathbf{x}_t$ is the noisy target feature at step $t$. This encourages the denoiser to reconstruct the original feature at the masked intermediate time point. This loss is also depicted as $L_{\text{Int}}$ in Fig.~\ref{fig:method}(a).

After initial training, the denoiser is used to augment data. For sequences with a single missing intermediate time point $i$, we use the available representations as the condition $\mathbf{C}$ and the target positional embedding $\mathbf{T}_i$ to impute the missing sMRI feature by sampling from the reverse diffusion process. These completed sequences are added to the training set, providing more complete longitudinal data.

\noindent \textbf{3.1.3 Extrapolation Task.} Extrapolation task focuses on training the denoiser to predict future sMRI representations beyond the observed sequence. This involves initial training and extrapolation-based data augmentation.
We train MCI-Diffusion to predict future sMRI features. For each complete sequence, we randomly choose a prediction horizon $k$ from $\{1, \ldots, |\mathcal{T}|-1\}$. To predict the sMRI feature at temporal index $i$, where $i \in \{|\mathcal{T}|-k, ..., |\mathcal{T}| - 1\}$, we mask not only the feature at $i$ but also all features at temporal indices later than $i$. The denoiser $\epsilon_\theta$ is conditioned on the resulting sequence $\mathbf{C}$ (with positions $i$ to $|\mathcal{T}|-1$ masked) and the target positional embedding $\mathbf{T}_i$. The training objective is to minimize the extrapolation loss:
\begin{equation}
L_{\text{Ext}} = \mathbb{E}_{\mathbf{\epsilon} \sim \mathcal{N}(0, 1), t} \left[ ||\mathbf{\epsilon} - \epsilon_\theta(\mathbf{x}_t, \mathbf{C}, \mathbf{T}_i, t)||_2^2 \right], \quad i \in \{|\mathcal{T}|-k, ..., |\mathcal{T}| - 1\}.
\end{equation}
where $\mathbf{x}_t$ is the noisy target feature at step $t$. This encourages the denoiser to reconstruct the original features at the masked future time points, relying only on information from time points before i. This loss is also depicted as $L_{\text{Ext}}$ in Fig.~\ref{fig:method}(a).

After initial training, we use the denoiser to augment data for extrapolation. We identify longitudinal sMRI sequences that are truncated, meaning they lack data, and crucially, \textbf{are missing only the final} time point. To mitigate error accumulation, we focus on predicting this single missing future time point. For each such sequence, we use the available \textbf{past} sMRI features as the condition $\mathbf{C}$ and sample from the reverse diffusion process, conditioned on $\mathbf{C}$ (where the final time point is masked) and the corresponding $\mathbf{T}_i$ for this missing \textbf{final} time point $i$, to generate the sMRI feature for that time point. These augmented sequences, now completed by predicting the last time point, are added to the training set, providing slightly more complete longitudinal data.

\noindent \textbf{3.1.4 Progressive Training Strategy.} To effectively train MCI-Diffusion with incomplete longitudinal sMRI sequences, we employ a progressive difficulty schedule. This strategy leverages both the Interpolation and Extrapolation tasks to maximize data utilization and learn robust representations. Algorithm~\ref{alg:progressive_training} provides a concise overview: the training starts with complete sequences and gradually incorporates incomplete sequences by imputing missing data using the model itself, while concurrently increasing the difficulty of the reconstruction task.

The progressive difficulty schedule is crucial. At the lowest difficulty, the model learns basic reconstruction (both interpolation and extrapolation of single missing points). As the difficulty increases, the model is challenged to make predictions based on less available information. At the maximum difficulty level ($d = D_{\text{max}}$) and in the Extrapolation task, the model predicts the subsequent sMRI feature conditioned solely on the baseline and any previously generated features, mirroring the initial step of autoregressive generation. This connection highlights how our training strategy seamlessly integrates the tasks and prepares the model for the ultimate goal of longitudinal prediction.

\subsection{LLM-driven Clinical Plausibility Sampling}

\begin{algorithm}[t]
\caption{LLM-guided Autoregressive Sampling}
\label{alg:LLM_guided_sampling}
\textbf{Input:} $\phi(X_0^{(p)})$, Diffusion Model, LLM, \textit{Tokenize}, \textbf{Instruction Template}, $T$, $N$\\
\textbf{Output:} $\{\hat{\mathbf{Z}}_\tau^{(p)}\}_{\tau=1}^{T}$

\begin{algorithmic}[1]
\For{$\tau = 1$ to $T$}
  \State $\{\hat{\mathbf{Z}}_\tau^{(p,n)}\}_{n=1}^{N} \leftarrow$ \textit{Diffusion Model} ($\phi(X_0^{(p)}), \{\hat{\mathbf{Z}}_{1:\tau-1}^{(p)}\}$)  \Comment{Generate candidates}
  \For{$n = 1$ to $N$}
    \State $\mathbf{t}_\tau^{(p,n)} \leftarrow$ \textit{Tokenize}($\hat{\mathbf{Z}}_\tau^{(p,n)}$)
    \State $\mathbf{y}_\tau^{(p,n)} \leftarrow$ LLM(\textbf{Instruction Template}, $\mathbf{t}_\tau^{(p,n)}$) \Comment{Predict FreeSurfer}
  \EndFor
  \State $\hat{\mathbf{Z}}_\tau^{(p)} \leftarrow \argmax_{\hat{\mathbf{Z}}_\tau^{(p,n)}} \textit{PlausibilityScore}(\{\mathbf{y}_\tau^{(p,n)}\}$) \Comment{Select best candidate}
\EndFor
\State \textbf{return} $\{\hat{\mathbf{Z}}_\tau^{(p)}\}_{\tau=1}^{T}$
\end{algorithmic}
\end{algorithm}

Following the training of our diffusion model via Multi-task Sequence Reconstruction, we introduce an LLM-driven approach to refine the sampling process and generate clinically plausible longitudinal sMRI feature sequences. This involves first instruction tuning the LLM to evaluate the clinical plausibility of these features (Instruction Tuning for Clinical Interpretation), and finally, integrating the LLM into the autoregressive sampling loop to guide the selection of the most clinically coherent feature sequences at each step (Autoregressive Sampling with LLM Guidance).

\noindent \textbf{3.2.1 Instruction Tuning for Clinical Interpretation.} 
To leverage the clinical knowledge embedded in LLMs, we first transform continuous sMRI features $\mathbf{Z}_\tau^{(p)} \in \mathcal{Z}$, generated by our model, into a discrete, sequential format suitable for LLM input by simply quantization and tokenization.

To imbue the LLM with the capability to provide clinically informed interpretations of generated sMRI representations, we perform instruction tuning focused on predicting interpretable structural features. 
To achieve this, we create a specialized dataset of paired tokenized sMRI features and their corresponding structural measurements, both extracted from the same set of sMRI scans within our training data. We then fine-tune the LLM using this dataset, providing instructions that guide it to predict the FreeSurfer measurements given the tokenized sMRI feature as input. An example of such an instruction template is shown in Fig.~\ref{fig: instruction}:
\begin{figure}[t]
  \centering
\includegraphics[width=\linewidth]{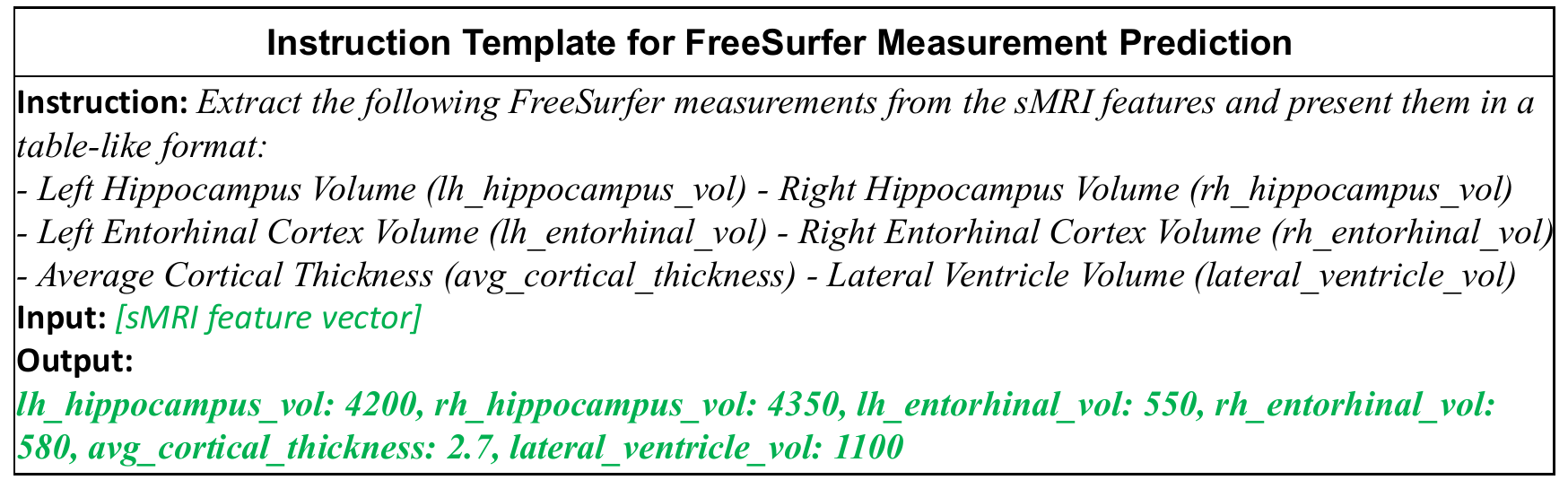}
\vspace{-5mm}
\caption{Illustration of an Instruction Template for structural Measurement Prediction.}
\vspace{-5mm}
   \label{fig: instruction}
\end{figure}
\noindent This instruction, paired with the corresponding tokenized sMRI feature vector as \textbf{Input} and the actual structural measurements as the \textbf{Output}, trains the LLM to learn the mapping between the abstract feature space and concrete clinical biomarkers.

\noindent \textbf{3.2.2 Autoregressive Sampling with LLM Guidance.}
Having adapted sMRI features into tokens and instruction-tuned the LLM to predict clinical measurements, we integrate the LLM into autoregressive sampling. At each step $\tau$, the diffusion model proposes $N$ candidate feature representations $\{\hat{\mathbf{Z}}_\tau^{(p,n)}\}_{n=1}^{N}$, which are tokenized $\{\mathbf{t}_\tau^{(p,n)}\}_{n=1}^{N}$ and fed to the LLM to obtain predicted measurements $\{\mathbf{y}_\tau^{(p,n)}\}_{n=1}^{N}$. The candidate with the most clinically plausible measurements (compared to expected values) is selected as $\hat{\mathbf{Z}}_\tau^{(p)}$ for the next step $\tau+1$. This iterative selection process, driven by the LLM's ability to interpret the generated features in terms of clinical biomarkers, guides the autoregressive sampling towards more realistic and clinically meaningful longitudinal trajectories.

\section{Experiment}

\subsection{Experimental Setup}
\noindent \textbf{Datasets.} Our method is evaluated on data from the Alzheimer's Disease Neuroimaging Initiative (ADNI) and the Australian Imaging Biomarkers and Lifestyle Study of Ageing (AIBL). Specifically, we utilize data from ADNI-1 and ADNI-2. These datasets include sMRI scans and corresponding demographic information for each participant, as summarized in the demographic table~\ref{tab:dataset_demographics}.

\noindent \textbf{Baselines.} The proposed model is compared with eight baseline models to evaluate its performance. For cross-sectional sMRI analysis, we select four established models: HFCN~\cite{HFCN}, DA-MIDL~\cite{DA-MIL}, MPS-FFA~\cite{MPS-FFA}, HMSA~\cite{HMSA}. 
For longitudinal sMRI analysis, we compare against two sequence-based models: AD-RNN~\cite{RNNMCI} and VGG-TS~\cite{2023vggts}. 
To provide a direct comparison between cross-sectional and longitudinal approaches, we adapt the prominent HFCN model to incorporate longitudinal information using a Transformer architecture, denoted as HFCN$^+$. 
Finally, for generative modeling, we include Temp-GAN~\cite{wang2018temporal} as a baseline for comparison.

\noindent \textbf{Evaluation Metrics.} The performance of our method was evaluated using four standard metrics: classification accuracy (ACC), which measures the overall correctness of the classifications; sensitivity (SEN), which quantifies the proportion of true positive cases correctly identified; specificity (SPE), which quantifies the proportion of true negative cases correctly identified; and the area under the receiver operating characteristic curve (AUC), which assesses the model's ability to discriminate between positive and negative cases across different thresholds.

\begin{wraptable}{h}{0.65\textwidth}
\centering
\caption{Demographics of ADNI and AIBL datasets.}
\label{tab:dataset_demographics}
\begin{tabular}{l c c c c}
\toprule
\textbf{Dataset} & \textbf{Group} & \textbf{Gender} & \textbf{Age} & \textbf{MMSE} \\
                 &                     & (M/F)           & (Mean$\pm$Std) & (Mean$\pm$Std) \\
\midrule
\multirow{2}{*}{ADNI-1}
& pMCI  & 102/65  & 74.78$\pm$6.85 & 26.59$\pm$1.71 \\
& sMCI  & 151/75  & 74.90$\pm$7.63 & 27.27$\pm$1.78 \\
\midrule
\multirow{2}{*}{ADNI-2}
& pMCI  & 24/14  & 71.34$\pm$7.30 & 27.02$\pm$1.69 \\
& sMCI  & 134/105  & 71.74$\pm$7.63 & 28.31$\pm$1.68 \\
\midrule
\multirow{2}{*}{AIBL}
& pMCI  & 9/8     & 75.29$\pm$6.16 & 26.24$\pm$2.04 \\
& sMCI  & 48/45   & 74.67$\pm$7.21 & 27.23$\pm$2.08 \\
\bottomrule
\end{tabular}
\end{wraptable}

\noindent \textbf{Implementation details.} In our experiments, we follow the established protocol~\cite{DA-MIL} of utilizing ADNI-1 data for training and ADNI-2 data for testing. For AIBL data, we use it to further verify the robustness and generalizability of our model. The sMRI images from these datasets were preprocessed by normalizing to MNI152 space using FSL, followed by skull stripping and bias field correction using ANTs. We use HFCN~\cite{HFCN} as the pretrained feature extractor to extract sMRI features and FreeSurfer 7.4.1 to extract structural features. We implement our model and associated components using PyTorch. Specific implementation details such as batch sizes, diffusion steps, and hardware configurations will be provided in the Appendix. All experiments are conducted on 5 NVIDIA V100 GPUs.

\subsection{Main Results}

\begin{table}[t]
  \centering
  \caption{Comparison of different methods on ADNI and AIBL datasets. Best results in \textbf{bold}.}
  \label{tab:main_performance}
  \sisetup{table-format=2.2} 
  \scalebox{0.98}{
  \begin{tabular}{ll|llll|llll}
    \toprule
    \multirow{2}{*}{\textbf{Category}} 
    & \multirow{2}{*}{\textbf{Method}} 
      & \multicolumn{4}{c|}{\textbf{ADNI}} 
      & \multicolumn{4}{c}{\textbf{AIBL}} \\
    \cmidrule(lr){3-6} \cmidrule(l){7-10}
    & 
      & \textbf{ACC} & \textbf{SEN} & \textbf{SPE} & \textbf{AUC}
      & \textbf{ACC} & \textbf{SEN} & \textbf{SPE} & \textbf{AUC} \\
    \midrule
    \multirow{4}{*}{\shortstack{\textbf{Cross-}\\\textbf{sectional}}}
      & HFCN      & 0.809 & 0.526 & 0.854 & 0.784 & 0.818 & 0.412 & 0.892 & 0.652 \\
      & DA‐MIDL   & 0.831 & 0.711 & 0.850 & 0.780 & 0.845 & 0.706 & 0.871 & 0.790 \\
      & MPS‐FFA   & 0.852 & 0.816 & 0.858 & 0.847 & 0.827 & 0.647 & 0.860 & 0.752 \\
      & HMSA      & 0.773 & 0.737 & 0.778 & 0.759 & 0.755 & 0.588 & 0.785 & 0.687 \\
    \midrule
    \multirow{3}{*}{\textbf{Longitudinal}}
      & AD‐RNN    & 0.751 & 0.684 & 0.762 & 0.720 & 0.755 & 0.471 & 0.806 & 0.639 \\
      & VGG‐TS    & 0.870 & 0.842 & 0.875 & 0.859 & 0.855 & 0.765 & 0.871 & 0.818 \\
      & HFCN$^+$ & 0.899 & 0.895 & 0.900 & 0.897 & 0.873 & 0.824 & 0.882 & 0.853 \\
    \midrule
    \multirow{3}{*}{\textbf{Generative}}
      & VAE           & 0.730 & 0.684 & 0.737 & 0.710 & 0.736 & 0.588 & 0.763 & 0.676 \\
      & Temp‐GAN  & 0.791 & 0.763 & 0.795 & 0.780 & 0.800 & 0.647 & 0.828 & 0.738 \\
      & \textbf{Ours} & \textbf{0.950} & \textbf{0.947} & \textbf{0.950} & \textbf{0.948} & \textbf{0.936} & \textbf{0.882} & \textbf{0.946} & \textbf{0.914} \\
    \midrule
    \multicolumn{2}{c|}{\textbf{Improvement (\%)}} & \textit{+5.1} & \textit{+5.2} & \textit{+5.0} & \textit{+5.1} & \textit{+6.3} & \textit{+5.8} & \textit{+6.4} & \textit{+9.6} \\
    \bottomrule
  \end{tabular}
  }
  \vspace{-5mm}
\end{table}

Table~\ref{tab:main_performance} shows our method outperforms baselines on ADNI and AIBL datasets across Accuracy, Sensitivity, Specificity, and AUC, with average improvements of 5.1-5.8\% (ADNI) and 5.8-11.7\% (AIBL). This indicates effective capture of cross-sectional and longitudinal information, surpassing cross-sectional models. While earlier longitudinal models underperform due to less sophisticated encoders, the improvement of HFCN+ over HFCN highlights the potential of longitudinal methods with effective encoding. Our method's outperformance of HFCN+ is attributed to multi-task learning for data completion, leading to more robust representations. The significant gains over generative models (VAE, Temporal-GAN) demonstrate our method's efficacy for discriminative tasks. Consistent improvements on both datasets confirm the robustness and generalizability of our approach.

\subsection{Abalation Study}

\begin{table}[ht]
\centering
\caption{Ablation Study on ADNI and AIBL Datasets}
\label{tab:ablation_study}
\scalebox{0.85}{ 
\begin{tabular}{l|l|cccc|cccc}
\toprule
\multirow{2}{*}{\textbf{Stage}} & \multirow{2}{*}{\textbf{Setting}} & \multicolumn{4}{c|}{\textbf{ADNI}} & \multicolumn{4}{c}{\textbf{AIBL}} \\
\cmidrule(lr){3-6} \cmidrule(lr){7-10}
& & ACC & SEN & SPE & AUC & ACC & SEN & SPE & AUC \\
\midrule
\multirow{4}{*}{\textbf{Training}}
& w/o Interpolation Task & 0.841 & 0.631 & 0.874 & 0.753 & 0.811 & 0.585 & 0.845 & 0.705 \\
& w/o Interpolation Aug. & 0.869 & 0.736 & 0.891 & 0.814 & 0.839 & 0.698 & 0.864 & 0.768 \\
& w/o Extrapolation Task & 0.838 & 0.737 & 0.853 & 0.803 & 0.820 & 0.647 & 0.849 & 0.748 \\
& w/o Extrapolation Aug. & 0.923 & 0.895 & 0.925 & 0.911 & 0.898 & 0.840 & 0.905 & 0.879 \\
\midrule
\multirow{2}{*}{\textbf{Sampling}}
& w/o Feature Adaptation & 0.893 & 0.842 & 0.899 & 0.870 & 0.865 & 0.795 & 0.875 & 0.838 \\
& w/o LLM-Guidance & 0.870 & 0.868 & 0.872 & 0.869 & 0.842 & 0.810 & 0.850 & 0.825 \\
\midrule
\multicolumn{2}{c|}{\textbf{Full Model (Ours)}} & \textbf{0.950} & \textbf{0.947} & \textbf{0.950} & \textbf{0.948} & \textbf{0.936} & \textbf{0.882} & \textbf{0.946} & \textbf{0.914} \\
\bottomrule
\end{tabular}
}

\end{table}

Ablation studies assessed component effectiveness: (1) w/o Interpolation Task; (2) w/o Interpolation Aug.; (3) w/o Extrapolation Task; (4) w/o Extrapolation Aug.; (5) w/o Feature Adaptation; (6) w/o LLM-Guidance; (7) Complete. Table \ref{tab:ablation_study} shows that removing the Extrapolation Task (3) substantially decreased performance on both datasets, highlighting its importance for future feature generation. Ablating LLM-Guidance (6) also reduced performance, indicating the value of clinical knowledge. The complete model (7) achieved the best performance, demonstrating each component's contribution.

\subsection{Hyperparameter Sensitivity Analysis}

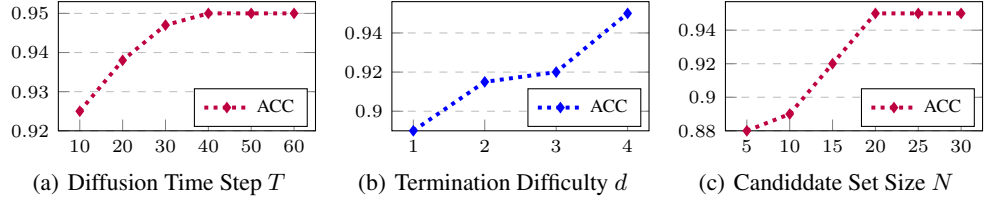
\begin{figure}[t!]
\centering

    \subfigure[Diffusion Time Step $T$]{\label{Fig-4a}
        \begin{tikzpicture}[font=\scriptsize]
            \begin{axis}[
                legend cell align={left},
                legend style={font=\scriptsize, nodes={scale=1.0, transform shape}},
                xlabel={},
                xtick pos=left,
                tick label style={font=\scriptsize},  
                xlabel style={font=\scriptsize}, 
                ylabel style={font=\scriptsize}, 
                ylabel={},
                ymin=0.92,
                width=5cm, 
                height=3.3cm, 
                xtick={10, 20, 30, 40, 50, 60},
                xticklabels={$10$, $20$, $30$, $40$, $50$, $60$},
                legend pos=south east,
                ymajorgrids=true,
                grid style=dashed,
            ]
            \addplot[
                color=purple,
                dotted,
                mark options={solid},
                mark=diamond*,
                line width=1.5pt,
                mark size=1pt
                ]
                coordinates {
                (10, 0.925)
                (20, 0.938)
                (30, 0.947)
                (40, 0.950)
                (50, 0.950)
                (60, 0.950)
                };
            \addlegendentry{ACC}
            
            \end{axis}
        \end{tikzpicture}
    }
    \subfigure[Termination Difficulty $d$]{\label{Fig-4b}
        \begin{tikzpicture}[font=\scriptsize]
            \begin{axis}[
                legend cell align={left},
                legend style={font=\scriptsize, nodes={scale=1.0, transform shape}},
                xlabel={},
                xtick pos=left,
                tick label style={font=\scriptsize},  
                xlabel style={font=\scriptsize}, 
                ylabel style={font=\scriptsize}, 
                ylabel={ },
                ymin=0.89,
                width=5cm, 
                height=3.3cm, 
                xtick={1, 2, 3, 4},
                xticklabels={$1$, $2$, $3$, $4$},
                legend pos=south east,
                ymajorgrids=true,
                grid style=dashed
            ]
            \addplot[
                color=blue,
                dotted,
                mark options={solid},
                mark=diamond*,
                line width=1.5pt,
                mark size=1pt
                ]
                coordinates { 
                (1, 0.890)
                (2, 0.915)
                (3, 0.920)
                (4, 0.950)
                };
            \addlegendentry{ACC}
            \end{axis}
        \end{tikzpicture}
    }
    \subfigure[Candiddate Set Size $N$]{\label{Fig-4c}
        \begin{tikzpicture}[font=\scriptsize]
            \begin{axis}[
                legend cell align={left},
                legend style={font=\scriptsize, nodes={scale=1.0, transform shape}},
                xlabel={},
                xtick pos=left,
                tick label style={font=\scriptsize},  
                xlabel style={font=\scriptsize}, 
                ylabel style={font=\scriptsize}, 
                ylabel={},
                ymin=0.88,
                width=5cm, 
                height=3.3cm, 
                xtick={5, 10, 15, 20, 25, 30},
                xticklabels={$5$, $10$, $15$, $20$, $25$, $30$},
                legend pos=south east,
                ymajorgrids=true,
                grid style=dashed,
            ]
            \addplot[
                color=purple,
                dotted,
                mark options={solid},
                mark=diamond*,
                line width=1.5pt,
                mark size=1pt
                ]
                coordinates { 
                (5, 0.88)
                (10, 0.89)
                (15, 0.92)
                (20, 0.95)
                (25, 0.95)
                (30, 0.95)
                };
            \addlegendentry{ACC}
            
            \end{axis}
        \end{tikzpicture}
    }
    \caption{Hyperparameter sensitivity results on ADNI.}
    \label{fig:para}
\end{figure}

\noindent \textbf{Diffusion Time Steps ($T$): } 
Fig.~\ref{fig:para} a illustrates the sensitivity of the model's accuracy to the diffusion time step ($T$). The results indicate that performance initially improves as $T$ increases, reaching a peak accuracy around T = 40. Beyond this point, the accuracy remains relatively stable, suggesting that increasing the number of diffusion steps beyond 40 does not significantly enhance the model's performance on the ADNI dataset.

\noindent \textbf{Termination Difficulty ($d$): } 
Fig.~\ref{fig:para} b shows the impact of the termination difficulty ($d$) on the model's accuracy. The plot reveals that accuracy generally increases with higher termination difficulty. The highest accuracy is achieved when $d = 4$, suggesting that a more challenging training regime leads to better performance. Lower termination difficulty might result in the model converging to a suboptimal solution, while a higher difficulty forces the model to learn more robust representations.

\noindent \textbf{Candidate Set Size ($N$): }  
Fig.~\ref{fig:para} c presents the analysis of the candidate set size ($N$), which ranges from 5 to 30. The accuracy demonstrates a clear upward trend as the candidate set size increases, with the most substantial gains observed up to $N = 20$. A larger candidate set likely provides the model with more diverse and potentially better options during sampling, leading to improved accuracy. However, after a certain point, the added diversity does not contribute further.

\subsection{Case Study}

\begin{figure}[t]
  \centering
\includegraphics[width=\linewidth]{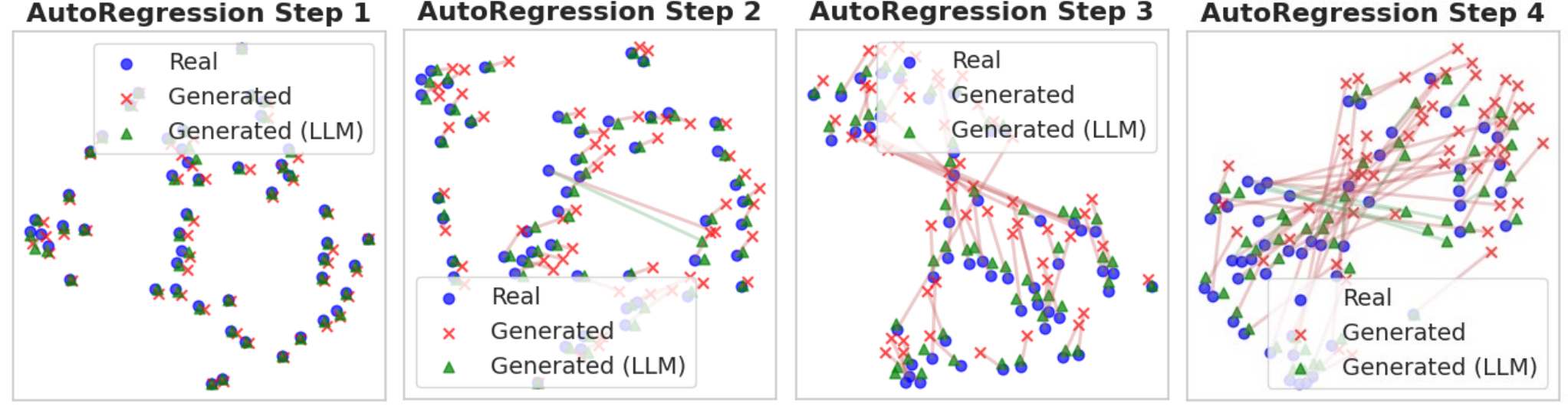}
\caption{Autoregressive generation of future MRI features over four steps, with and without LLM-driven Clinical Plausibility Sampling.}
\label{fig: case}
\end{figure}
To qualitatively assess our framework, we present a case study visualizing the four-step autoregressive generation of future MRI representations conditioned on baseline data, comparing uncorrected diffusion model outputs (red crosses) with real future features (blue circles) and LLM-refined features (green triangles). 
As the generation progresses, uncorrected features diverge, indicating error accumulation, while LLM-corrected features consistently align more closely with the real data distribution, demonstrating the LLM's effectiveness as a "linguistic compass" in steering the generation towards clinically plausible trajectories and mitigating instability inherent in sequential diffusion modeling. 
This visual evidence underscores the crucial role of LLM integration in producing more reliable future MRI feature predictions for early AD detection.

\section{Conclusion}

We present \textbf{MCI-Diff}, a diffusion framework that synthesizes future sMRI features from a single baseline scan and refines them with LLM-guided plausibility scoring. Our multi-task denoising learns robust latent trajectories, while quantization, tokenization, and LLM evaluation ensure clinically realistic outputs. On ADNI and AIBL data, MCI-Diff outperforms existing methods, improving early MCI conversion accuracy by up to 12 \%. This fusion of generative modeling and language-driven clinical insight paves the way for accurate, real-time neurodegenerative prognosis.  

\bibliographystyle{plain}

\bibliography{neurips_2025}




\end{document}